%% file: ms.tex
\documentclass[sigconf,authorversion, nonacm]{acmart}
\usepackage[utf8]{inputenc}
\usepackage{tikz,graphicx}

\usepackage{algorithm}
\usepackage{algorithmic}
\usepackage{amsmath}
\usepackage{booktabs}
\usepackage{xspace}
\usepackage{subcaption}
\usepackage{hyperref}
\usepackage{listings}
\usepackage{tabularx}
\usepackage{nicefrac}

\lstdefinelanguage{ttl} {
language={SQL},
    alsoletter={-},
    morekeywords={geovec, geovec-s, a, lgd, geo, dcterms, rdfs, owl, prov, xsd},
}

\usepackage{stmaryrd} 

\newcommand{\voc}[2]{\texttt{#1:\allowbreak #2}}

\title{GeoVectors: A Linked Open Corpus of OpenStreetMap Embeddings on World Scale}

 \author{Nicolas Tempelmeier}
 \affiliation{%
   \institution{L3S Research Center\\Leibniz Universit\"at Hannover}
   \city{Hannover}
   \country{Germany}
 } \email{tempelmeier@L3S.de}

 \author{Simon Gottschalk}
 \affiliation{%
   \institution{L3S Research Center\\Leibniz Universit\"at Hannover}
   \city{Hannover}
   \country{Germany}
 } \email{gottschalk@L3S.de}

 \author{Elena Demidova}
 \affiliation{%
   \institution{\mbox{Data Science \& Intelligent Systems (DSIS)} University of Bonn}
   \city{Bonn}
   \country{Germany}}
 \email{elena.demidova@cs.uni-bonn.de}

\renewcommand{\shortauthors}{N. Tempelmeier, S. Gottschalk, and E. Demidova}

\newcommand{\corpus}{GeoVectors\xspace}
\newcommand{\gvtags}{\textit{GV-Tags}\xspace}
\newcommand{\gvnle}{\textit{GV-NLE}\xspace}

\begin{document}


\input{0_abstract}

\maketitle

\noindent\textbf{Resource type:} Dataset\\
\textbf{Documentation:} \url{http://geovectors.l3s.uni-hannover.de} \\
\textbf{Dataset DOI:} \url{https://doi.org/10.5281/zenodo.4339523} \\

\input{1_Introduction}

\input{2_impact}

\input{3_Generation}

\input{5_LOD-Int}

\input{4_Characteristics}

\input{6_Case_Study}

\input{7_Availability}

\input{8_RelatedWork}

\input{9_Conclusion}

\subsubsection*{Acknowledgements} This work was partially funded by DFG, German Research Foundation (``WorldKG'', 424985896), the Federal Ministry of Education and Research (BMBF), Germany (``Simple-ML'', 01IS18054), the Federal Ministry for Economic Affairs and Energy (BMWi), Germany (``d-E-mand'', 01ME19009B), and the European Commission (EU H2020, ``smashHit'', grant-ID 871477).

\balance
\bibliographystyle{ACM-Reference-Format}
\bibliography{ref}

\end{document}

%% file: 0_abstract.tex
\begin{abstract}
OpenStreetMap (OSM) is currently the richest publicly available information source on geographic entities (e.g., buildings and roads) worldwide. However, using OSM entities in machine learning models and other applications is challenging due to the large scale of OSM, the extreme heterogeneity of entity annotations, and a lack of a well-defined ontology to describe entity semantics and properties.
This paper presents GeoVectors -- a unique, comprehensive world-scale linked open corpus of OSM entity embeddings covering the entire OSM dataset and providing latent representations of over 980 million geographic entities in 180 countries. 
The GeoVectors corpus captures semantic and geographic dimensions of OSM entities and makes these entities directly accessible to machine learning algorithms and semantic applications. 
We create a semantic description of the GeoVectors corpus, including identity links to the Wikidata and DBpedia knowledge graphs to supply context information. Furthermore, we provide a SPARQL endpoint -- a semantic interface that offers direct access to the semantic and latent representations of geographic entities in OSM.  
\end{abstract}

%% file: 1_Introduction.tex
\section{Introduction}
\label{sec:intro}

OpenStreetMap (OSM) has evolved as a critical source of openly available volunteered geographic information \cite{JokarArsanjani2015}.
The amount of information available in OpenStreetMap is continuously growing. 
For instance, the number of geographic entities captured by OSM increased from $5.9 \cdot 10^9$ in March 2020 to $6.7 \cdot 10^9$ in March 2021.
Today, OSM data is used in a plethora of machine learning applications such as road traffic analysis \cite{Keller_2020}, remote sensing \cite{9119753}, and geographic entity disambiguation \cite{TEMPELMEIER2021349}.
Other data-driven OSM applications include map tile generation \cite{4653466} and routing \cite{epub34045}.

OpenStreetMap is a collaborative online project that aims to create a free and editable map and features over 7.6 million contributors as of June 2021.
OSM provides information about its entities in the form of key-value pairs, so-called \emph{tags}. For instance, the tag \texttt{place=city} indicates that an entity represents a city.
OSM tags and keys do not follow any well-defined ontology or controlled vocabulary. 
Instead, OSM encourages its contributors to follow a set of best practices for annotation\footnote{\url{https://wiki.openstreetmap.org/wiki/Any\_tags\_you\_like}}. 
The number of tags and the level of detail for individual OSM entities is highly heterogeneous \cite{Touya2015}.
For instance, as of June 2021, the size of data available for the country of Germany sums up to 3.3 GB, while only 2.5 GB of data is available for the entire South American continent.
Ultimately, factors including 1) a varying number of tags and details for specific geographic entities, 2) the lack of a well-defined ontology resulting in numerous tags with unclear semantics, and 3) missing values for any given property, substantially hinder the feature extraction for broader OSM usage in machine learning applications.

A central prerequisite to facilitate the effective and efficient use of geographic data in machine learning models is the availability of suitable representations of geographic entities.
Recently, latent representations (embeddings) have been shown to have several advantages in machine learning applications, compared to traditional feature engineering, in a variety of domains \cite{LIU2018183,Xie:2016:LGP:2983323.2983711,Wang:2017:RRL:3132847.3133006}. 
First, embeddings can capture semantic entity similarity not explicitly represented in the data.
Second, embeddings facilitate a compact representation of entity characteristics, overall resulting in a significant reduction of memory consumption \cite{TEMPELMEIER2021349}. 
Whereas much work has been performed to provide pre-trained embeddings for textual data and knowledge graphs \cite{DBLP:journals/computing/WangZJ20,8047276}, only a few attempts, such as \cite{10.1007/978-3-319-68204-4_14}, aimed to provide such latent representations for geographic entities and captured selected entities only.
From the technical perspective, the creation of OSM embeddings is particularly challenging due to the large scale of OSM (more than 1430 GB of data as of June 2021\footnote{\url{https://wiki.openstreetmap.org/wiki/Planet.osm}}) and the OSM data format (\emph{``protocolbuffer binary format''}\footnote{\url{https://wiki.openstreetmap.org/wiki/PBF_Format}}), requiring powerful computational infrastructure and dedicated data extraction procedures.
Furthermore, the semi-structured data format of OSM tags requires specialized embedding algorithms to capture the semantics of entity descriptions.
As a result of these challenges, currently, no datasets that capture latent representations of OSM entities exist. 

The \corpus{} corpus of embeddings presented in this paper is a significant step to enable the efficient use of extensive geographic data in OSM by machine learning algorithms. \corpus facilitates access to these embeddings using semantic technologies. 
We utilize established representation learning techniques (word embeddings and geographic representation learning) to capture various aspects of OSM data. 
We demonstrate the utility of the \corpus{} corpus in two case studies covering 
the tasks of type assertion and link prediction in knowledge graphs.
\corpus{} follows the \emph{5-Star Open Data} best practices in data publishing and reuses existing vocabularies to lift OpenStreetMap entities into a semantic representation. 
We provide a knowledge graph that semantically represents the \corpus{} entities and interlinks them with existing resources such as Wikidata, DBpedia, and Wikipedia.
With the provision of pre-computed latent OSM representations, we aim to substantially ease the use of OSM entities for machine learning algorithms and other applications.

To the best of our knowledge, currently, there are no dedicated resources that provide extensive reusable embeddings for geographic entities at a scale comparable to \corpus.
The absence of comprehensive geographic data following a strict schema makes it particularly challenging to process geographic data in machine learning environments. 
We address these problems by providing models capable of embedding arbitrary geographic entities in OSM. 
Moreover, we enable easy reuse by making both models and encoded data publicly available.

The main contributions of this paper are as follows:
\begin{itemize}
    \item We provide \corpus{} -- a world-scale corpus of embeddings covering over 980 million geographic entities in 188 countries using two embedding models and capturing the semantic and the geographic dimensions of OSM entities.
    \item We introduce an open-source embedding framework for OSM to facilitate the reusable embedding of up-to-date entity representations\footnote{\url{https://github.com/NicolasTe/GeoVectors}}.
    \item We provide a knowledge graph to enable semantic access to \corpus{}.
\end{itemize}
The remainder of this paper is organized as follows. 
In Section ~\ref{seC:relevance}, we discuss the predicted impact of \corpus.
Then, in Section ~\ref{sec:generation}, we present the embedding generation framework.
In Section ~\ref{sec:kg}, we present the \corpus knowledge graph.
Next, we describe the characteristics of the \corpus corpus in Section ~\ref{sec:chararcteristics}.
We illustrate the usefulness of \corpus in two case studies in Section ~\ref{sec:case_study} and discuss availability and utility in Section ~\ref{sec:avail}.
In Section ~\ref{sec:related_work}, we discuss related work.
Finally, in Section  ~\ref{sec:conclusion}, we provide a conclusion.

%% file: 2_impact.tex
\section{Predicted Impact}
\label{seC:relevance}

\corpus is a new resource. 
This section discusses the predicted impact of \corpus regarding the advances of state of the art in geographic embedding datasets, geographic information retrieval, machine learning applications, knowledge graph embeddings
and broader adoption of semantic web technologies.
\textit{Advances of the state of the art:}
We advance the state of the art by providing the first large-scale corpus of pre-trained geographic embeddings.
We carefully select established representation learning techniques to capture both the semantic dimension (What entity type does the OSM entity represent?) and the spatial dimension (Where is the entity located?) and adapt these techniques to OSM data to create meaningful latent representations.
The \corpus corpus is the first dataset that captures the entire OpenStreetMap, thus offering the data on the world scale.
Therefore, \corpus is significantly larger than any existing geographic embedding resources. For instance, the Geonames embedding \cite{10.1007/978-3-319-68204-4_14} provides a dataset containing less than 358 thousand entities, whereas \corpus contains over 980 million entities.

\textit{Impact on geographic information retrieval:}
Geographic information retrieval (GIR) addresses the problems of developing location-aware search systems and addressing geographic information needs \cite{INR-034}.
Recent GIR approaches build on geographic embeddings to address several use cases, including tag recommendation for urban complaint management \cite{10.1145/3357384.3357894},  geographic question answering \cite{10.1145/3442381.3449857}, and POI categorization \cite{10.1145/3397536.3422196}.
While these approaches demonstrate the utility of geographic embeddings for GIR tasks, the laborious generation process hinders the adaption of geographic embeddings for other GIR tasks such as geographic named entity recognition, next location recommendation, or geographic relevance ranking.
In this context, the availability of large-scale and accessible geographic embeddings is a vital prerequisite to stimulate research in the GIR field.
The \corpus corpus presented in this paper addresses these requirements by providing ready-to-use geographic embeddings of the entire OpenStreetMap.

\textit{Impact on machine learning applications}:
Existing machine learning applications use geographic data to address numerous use cases including location recommendation \cite{LIU2018183,Xie:2016:LGP:2983323.2983711}, human mobility prediction \cite{Wang:2017:RRL:3132847.3133006}, and travel time estimation \cite{10.1145/3424346}.
The variety of use cases highlights the general importance of geographic information for machine learning models. 
However, these approaches conduct a costly feature extraction process or learn supervised embeddings of geographic entities on task-specific datasets for specific tasks. 
In this context, the availability of easy-to-use representations of geographic entities at scale provided by \corpus is crucial to enabling and easing the further development of geographic machine learning models and geographic algorithms.

\textit{Impact on knowledge graph embeddings:}
Knowledge graph embeddings generated without the specific focus on geographic entities have shown success in a large variety of knowledge graph inference and enrichment tasks, including type assertions and link prediction \cite{paulheim2017knowledge}. We envision that \corpus{} can further enhance the quality of embeddings used in the context of these tasks: While geographic entities are part of many popular knowledge graphs such as Wikidata and DBpedia, their specific characteristics are still rarely considered. Existing approaches typically focus on the graph structure, but rarely on the property values assigned to the single nodes \cite{kristiadi2019incorporating}. However, both tags and coordinates of geographic entities bear valuable semantics. 
Specifically, the geographic interpretation of coordinates may heavily lift the role of coordinates in knowledge graph embeddings.
In the future, the \corpus{} embeddings can directly support knowledge graph inference and enrichment and creation of geographically aware embeddings from other sources.

\textit{Impact on adoption of semantic web technologies:}
In the context of the Semantic Web, a variety of models and applications, including link prediction, creation of domain-specific knowledge graphs \cite{gottschalk2019eventkg} and Question Answering for event-centric questions \cite{CostaGD20} make use of geographic data.
Semantic technologies have been applied to a variety of domains that require spatio-temporal data, including crime localization, transport data, and historical maps \cite{10.1007/978-3-030-62466-8_23,10.1007/978-3-030-62466-8_26,10.1007/978-3-030-49461-2_24}. 
Furthermore, with the increased availability of mobile devices, location-based algorithms such as next location recommendation or trip planning evolved.
Recently, SPARQL extensions for integrated querying of semantic and geographic data have been proposed \cite{10.1007/978-3-030-62419-4_15}.
In this context, the availability of easy-to-use representations of geographic entities at scale is crucial to enable further development of semantic models and geographic algorithms and their adoption in real-world scenarios.
Increasing availability of geographic data accessible through semantic technologies, as facilitated by \corpus, and seamless integration of this data with other semantic data sources in the Linked Data Cloud can attract interested users from various disciplines and application domains, including geography, mobility, and smart cities.

%% file: 3_Generation.tex
\section{\corpus Framework for Embedding Generation}
\label{sec:generation}

\begin{figure*}[t!]
    \centering
    \includegraphics[width=0.7\textwidth]{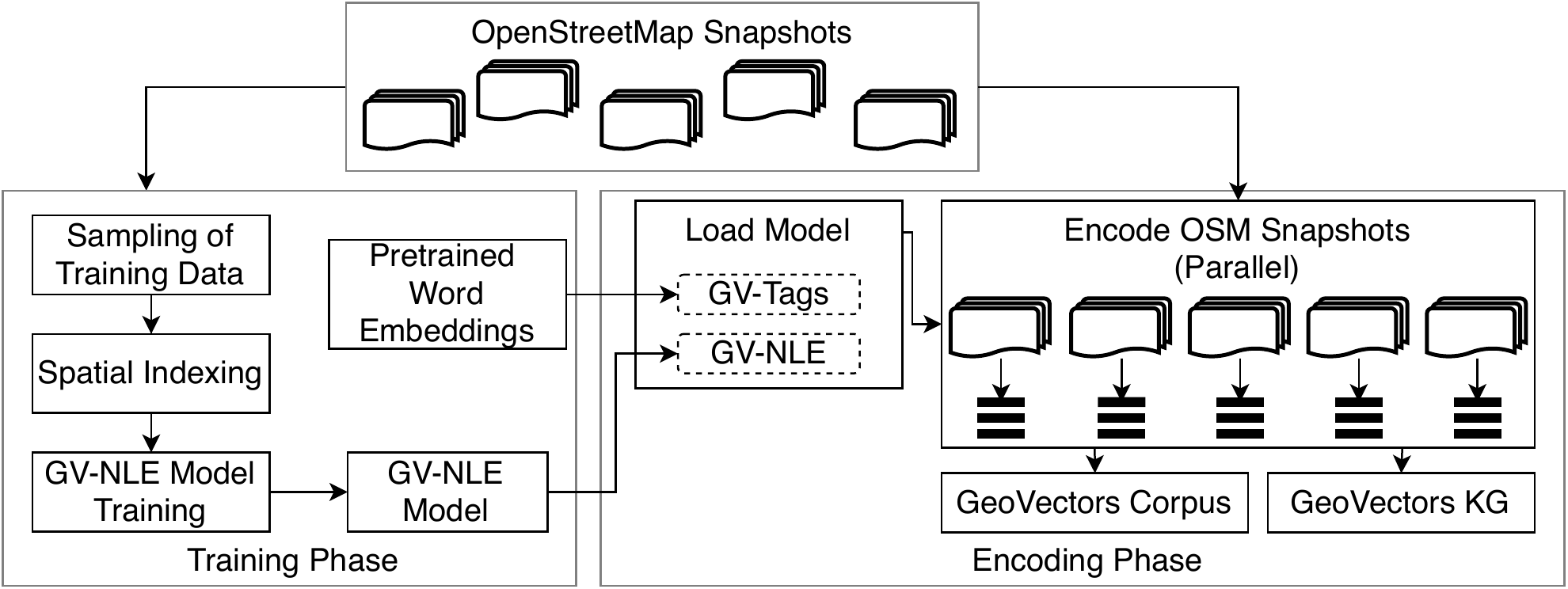}
    \caption{Overview of the embedding generation process.}
    \label{fig:genereation_overview}
\end{figure*}

The \corpus framework facilitates the generation of OSM embeddings that capture geographical (\gvnle{}) and semantic (\gvtags{}) similarity of OSM entities.
In this section, we first describe the OSM data model. Then, we provide an overview of the \corpus embedding generation process and 
present embedding algorithms that generate the proposed \gvnle{} and \gvtags{} embeddings.  

\subsection{OpenStreetMap Data Model}
\label{sec:osm-data-model}

The OSM data model distinguishes three entity types: \emph{nodes}, \emph{ways} and \emph{relations}.

\begin{itemize}
    \item \emph{Nodes} represent geographic points. A pair of geographic coordinates gives the point location. Examples of nodes include city centers and mountain peaks.
    
    \item \emph{Ways} represent entities in the form of a line string, e.g., roads. Ways aggregate multiple nodes that describe the pairwise segments of the line string.
    
    \item \emph{Relations} represent all other aggregated geographic entities. Relations consist of a collection of arbitrary OSM entities to form, for instance, a polygon. Examples of relations include country boundaries and train routes.
\end{itemize}

Each OSM entity can have an arbitrary number of annotations called \emph{tags}. Tags are key-value pairs that describe entity features. For example, the tag \texttt{place = city} indicates that a particular OSM entity represents a city.
More formally, an \emph{OSM entity} $o = \langle id, type, T \rangle$ consists of an identifier $id$, a $type \in \{\textit{node, way, relation}\}$, and a set of tags $T$. An \emph{OSM snapshot} taken at a time $t$ in a region $r$ is denoted by $s = \langle O, t, r \rangle$, where $O$ is a set of OSM entities within the specified region $r$ at this time.

\subsection{\corpus Embedding Generation Overview}
\label{sec:embedding}

The \corpus{} embeddings reflect semantic and geographic relations of OSM entities, where 
semantic relations capture semantic entity similarity, expressed through shared annotations, and geographic relations capture geographic entity proximity. 
In general, the relevant relation type is application-dependent. 
Therefore, we compute two embedding datasets, one capturing geographic and the other semantic similarity of OSM entities: 
\begin{itemize}
    \item (1) \emph{\gvnle{}} is our geographic embedding model based on the Neural Location Embeddings (NLE) \cite{10.1007/978-3-319-68204-4_14} -- an approach to capture the spatial relations of geographic entities.
    \item (2) \emph{\gvtags{}} is our semantic embedding model based on fastText \cite{joulin-etal-2017-bag} -- a state-of-the-art word embedding model that we apply on the OSM tags.
\end{itemize}

The embedding generation process that takes as input a set of OSM snapshots and generates the \corpus{} corpus and the \corpus{} knowledge graph is illustrated in Figure \ref{fig:genereation_overview}. 
We divide this process into the \emph{training} phase in which we train the GV-NLE model and the \emph{encoding} phase in which we apply embedding models to encode OSM entities.

The training of an embedding model is typically significantly more expensive than the application of the model. 
Due to the large scale of OpenStreetMap (as of June 2021, OSM contains more than 7 billion entities), the training of embedding models on the entire corpus is not feasible.
Therefore, in the \emph{training} phase, we sample a subset of OSM entities from OSM snapshots to serve as training data. 
We discuss the sampling process in Section \ref{sec:sampling}.
Based on the sampled data, we train our embedding models.
To generate semantic embeddings, we utilize existing pre-trained word embedding models. 

In the \emph{encoding} phase, we first load the trained embedding model and then pass all individual entities from an OSM snapshot to the model.
The application of the model can be parallelized by applying the model to each snapshot separately.
The model encodes the OSM entities and stores the generated embedding vectors into an easily processable, tab-separated value file.

We provide an open-source implementation of the embedding framework, including the pre-trained embedding models\footnote{\url{https://github.com/NicolasTe/GeoVectors}}. This framework enables the computation of up-to-date embeddings of individual OpenStreetMap snapshots. We also generate the \corpus knowledge graph that enables semantic access to \corpus and is described in detail in Section \ref{sec:kg}.

We performed the entire extraction and embedding process on a server with 6 TB of memory and 80 Intel(R) Xeon(R) Gold 5215M 2.50GHz CPU cores. Our framework required about four days for data extraction, model training, and  data encoding.

\subsection{Sampling of OSM Training Data for Embedding Algorithms}
\label{sec:sampling}

At the beginning of the training phase, we extract a representative entity subset to use as a training set. 
To ensure representativeness, we employ the following conditions:
First, the training set should have a balanced geographic distribution to avoid biases towards specific geographic regions.
Second, the training set should only include meaningful OSM entities. 
For instance, many OSM nodes do not provide any tags and only represent spatial primitives for composite entities, such as ways and relations. Such nodes do not correspond to real-world entities and, taken isolated, do not convey any meaningful information. Therefore, we exclude nodes without tags from the training data.

\begin{algorithm}
\begin{flushleft}
\begin{tabular}{lll}
     \texttt{Input:} & $\mathcal{S}$ & OpenStreetMap snapshots  \\
                     & $n$ & Minimum number of training examples \\
     \texttt{Output:} & $\mathcal{R}$ & Set of training examples\\
\end{tabular}
\end{flushleft}
\hrule
\begin{algorithmic}[1]
    \STATE \text{total\_area} $ \leftarrow$ $\sum_{s \in S}\texttt{geo\_area}(s.r)$
    
    \STATE $\mathcal{R} \leftarrow \{\}$
    \FORALL{$s \in \mathcal{S}$}
        \STATE $n_s \leftarrow n ~\cdot$ \texttt{geo\_area}(s.r) / total\_area 
        \STATE linked, tagged, other $\leftarrow$ \texttt{scan\_snapshot}($s$)
        \STATE $\mathcal{T} \leftarrow$ linked
        \IF {$|\mathcal{T}| < n_s$}
             \STATE $\mathcal{T} \leftarrow \mathcal{T} ~\cup$ \texttt{sample}(tagged, $(n_s - |\mathcal{T}|)$)
        \ENDIF
        \IF {$|\mathcal{T}| < n_s$}
             \STATE $\mathcal{T} \leftarrow \mathcal{T} ~\cup$ \texttt{sample}(other, $(n_s - |\mathcal{T}|)$)
        \ENDIF
        \STATE $\mathcal{R} \leftarrow \mathcal{R} \cup \mathcal{T}$
    \ENDFOR
    \RETURN $\mathcal{R}$
\end{algorithmic}
\caption{Sample Training Data}
\label{alg:Sample}
\end{algorithm}

Algorithm \ref{alg:Sample} presents the sampling process to obtain training data.
The input of the algorithm consists of a minimum number $n$ of training samples to be collected and a corpus of OpenStreetMap snapshots $\mathcal{S}$ (e.g., country-specific snapshots). 
First, we calculate the total geographic area covered by all snapshots using the $\texttt{geo\_area}(s.r)$ function (line 1), where $s.r$ denotes the region of the OSM snapshot $s$.
To enforce a uniform geographic distribution, we calculate the number of samples extracted from a single snapshot regarding its geographic size.
For each snapshot, we determine the number of samples $n_s$ to be extracted proportionally to the geographic area of the snapshot (line 4).
Then, the \texttt{scan\_snapshot} function divides the snapshot into \emph{linked}, \emph{tagged} and \emph{other} entities (line 5).
Linked entities provide an identity link to external datasets. 
As identity links typically indicate good data quality, our algorithm includes all linked entities.
Tagged entities provide at least one tag.
Other entities are entities that neither provide an identity link nor a tag.
Next, the algorithm samples all linked entities (even if their number exceeds $n_s$) into the result set $\mathcal{T}$ (line 6).
If the size of $\mathcal{T}$ does not reach $n_s$, the function \texttt{sample} uniformly selects at maximum $n_s -|\mathcal{T}|$ random samples from the tagged entities (lines 7-9).
If $n_s$ is still not reached, we sample the remaining examples from the other entities (lines 10-12).
Finally, the algorithm returns the union of all snapshot-specific training examples $\mathcal{R}$ (lines 13-15).

\subsection{\gvnle{} Embedding of OSM Entity Locations}
\label{sec:neural-embed}

The GV-NLE model builds on the neural location embedding (NLE) model \cite{10.1007/978-3-319-68204-4_14} that captures the geographic relations of a set of geographic entities in a latent representation.
The NLE method is an established method to create reusable geographic embeddings.
GV-NLE extends the NLE model with a suitable encoding algorithm to encode previously unseen OSM entities.

\emph{Training:} GV-NLE first constructs a weighted graph representing OSM entities and their mutual distances. 
The OSM entities form the nodes of the graph.
The edges encode the geographic distance between OSM entities. 
For each node $n$, GV-NLE constructs edges between $n$ and the $k$ geographically nearest neighbor nodes.
Following \cite{10.1007/978-3-319-68204-4_14}, we set $k=50$.
The edge weights represent the haversine distance between two nodes in meters, which measures the geographic distance of two points while taking the earth's curvature into account.
To facilitate an effective distance computation between OSM entities, we employ a Postgres database that provides spatial indexes.
Based on the graph, a weighted DeepWalk algorithm \cite{Perozzi:2014:DOL:2623330.2623732} learns the latent representations of the OSM nodes.
GV-NLE computes a damped weight $w' = \textit{max}(\nicefrac{1}{ln(w)}, e)$, where $w$ denotes the original edge weight, $ln$ the natural logarithm, and $e$ Euler's number.
The use of damped weights further prioritizes short distances between the nodes.
The normalized damped weights serve as a probability distribution for the transition probabilities of the random walk within the DeepWalk algorithm.

\emph{Encoding:} As the original NLE algorithm does not generalize to unseen entities, i.e., entities that are not part of the training set, we extend the NLE model with a suitable encoding algorithm.
The idea of the GV-NLE encoding is to infer a representation of an entity from its geographically nearest neighbors. 
We calculate the weighted average of the latent representation of the geographically nearest $k=50$ entities in the training set.

\[
    \textit{GV-NLE}(o) = \frac{1}{\sum_{o' \in N_o} w(o, o')}\sum_{o' \in N_o} w(o, o') \cdot \textit{NLE}(o')
\]
Here, $o$ denotes an OSM entity, $\textit{NLE}(o')$ denotes the latent representation of an entity $o'$ according to the NLE algorithm, $N_o$ denotes the set of the $k$ geographically nearest OSM entities in the training set.
We define the weighting term $w(o, o')$ as
\[
 w(o, o') = ln(1 + \frac{1}{dist(o, o')})
\]
where $dist(o, o')$ denotes the geographic distance between $o$ and $o'$.
$w(o,o')$ assigns a higher weight to geographically closer entities.
We apply a logarithm function to soften high weights of very close entities.

\subsection{\gvtags{} Embedding of OSM Entity Tags}
\label{sec:fastText}

To infer the \gvtags{} representations, we adopt fastText, a state-of-the-art word embedding model that infers the latent representation of single words individually \cite{joulin-etal-2017-bag}. 
As the tags of OSM entities do not have any natural order, we chose fastText to embed them.

\emph{Training:} Pre-trained word vectors are available at the fastText website\footnote{\url{https://fastText.cc/}}. 
As most of the OSM keys are in English, we chose the 300-dimensional English word vectors trained on the Common Crawl, and Wikipedia \cite{grave2018learning}.

\emph{Encoding:}
To encode an OSM entity $o$, we utilize the individual word embeddings of the keys and values that form the entity tags $o.T$. We map entities without any tags to a vector of zeros.

\[
 \gvtags{}(o) =
\begin{cases}
        \frac{1}{2|o.T|} \cdot \sum_{\langle k,v \rangle \in o.T} ft(k) + ft(v), &  if ~| o.T | > 0\\
        \{0\}^{300},      & \text{otherwise}.
\end{cases} 
\]

Here, $\{0\}^{300}$ denotes a 300-dimensional vector of zeros, and $ft(x)$ denotes the fastText word embedding of $x$.

%% file: 5_LOD-Int.tex
\section{GeoVectors Knowledge Graph}
\label{sec:kg}

\begin{figure*}[t!]
    \centering
    \includegraphics[width=0.7\textwidth]{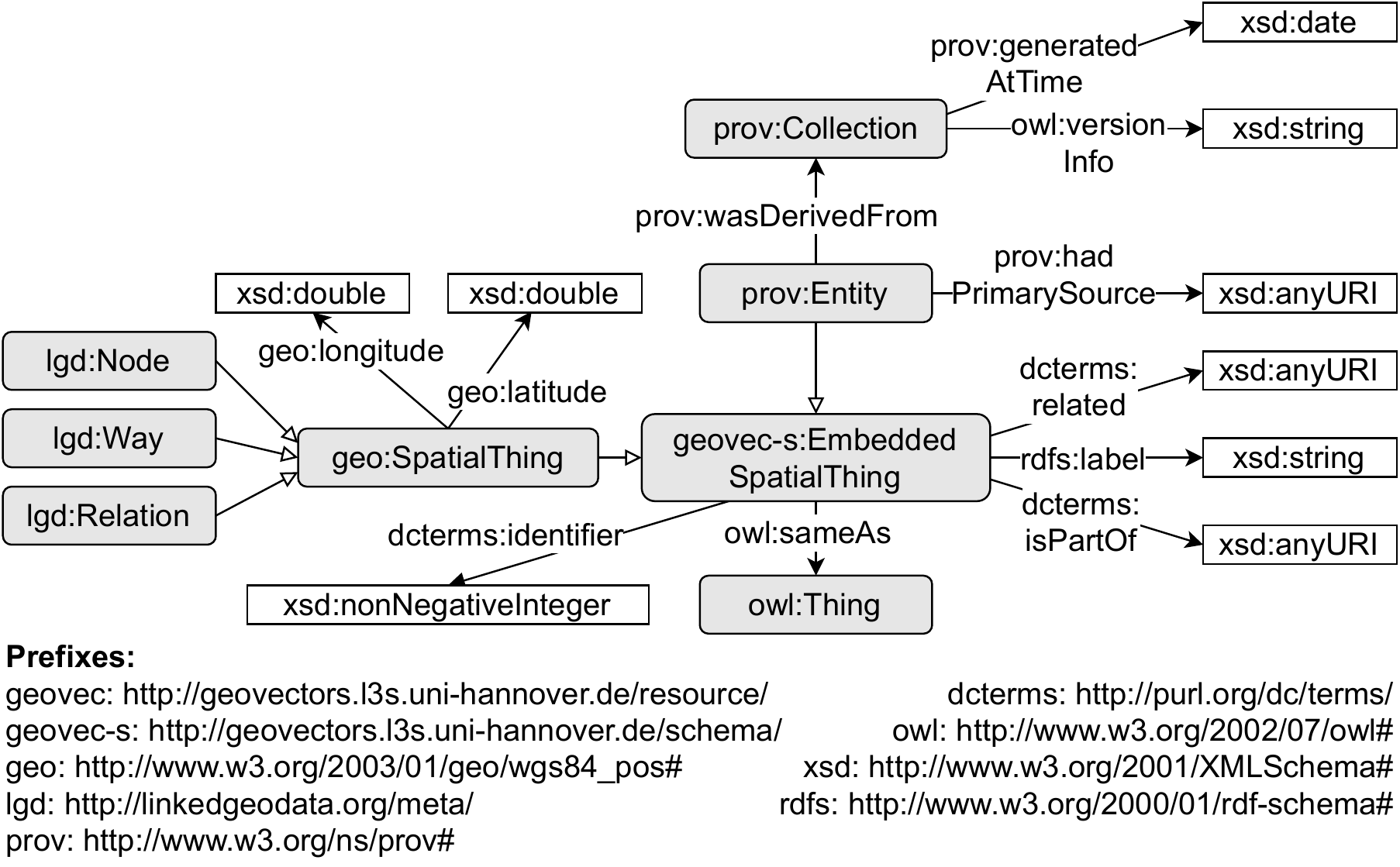}
    \caption{Schema, prefixes and namespaces of the \corpus{} knowledge graph. $\rightarrowtriangle$ marks a \voc{rdfs}{subClassOf} relation, \linebreak $\rightarrow$ denotes the domain and range of a property.}
    \label{fig:rdf_schema}
\end{figure*}

Semantic access to the \corpus{} embeddings is of utmost importance to facilitate the use of the dataset in downstream semantic applications. 
Therefore, \corpus{} is accompanied by a knowledge graph that models the embedding metadata. 
This metadata facilitates interlinking of the embeddings with established knowledge graphs such as Wikidata and DBpedia using existing entity links. 
This way, the \corpus{} embeddings can be used to enrich geographic entities in these knowledge graphs.
The \corpus{} knowledge graph includes more than $28$ million triples and is made available under a public SPARQL endpoint\footnote{\url{http://geovectors.l3s.uni-hannover.de/sparql}}. 

The \corpus{} knowledge graph is based on three established vocabularies. We utilize the LinkedGeoData \cite{10.5555/2590208.2590210} and the Basic Geo vocabulary\footnote{\url{https://www.w3.org/2003/01/geo/wgs84_pos}} to model the spatial dimension of geographic entities, as well as the PROV Ontology \cite{lebo2013prov} for modeling data provenance, i.e., where the geographic entities were extracted from and what they represent. Figure \ref{fig:rdf_schema} illustrates the schema of the \corpus{} knowledge graph, including its prefixes and namespaces.

Each geographic entity in the knowledge graph is typed as \voc{geovec}{Embedded\-Spatial\-Thing}, which encapsulates the classes \voc{geo}{Spatial\-Thing} and \voc{prov}{En\-ti\-ty}. We group the relevant properties shown in Figure \ref{fig:rdf_schema} regarding these three classes:

\begin{itemize}
\item \voc{geo}{SpatialThing}: Each geographic entity is either a node, a way or a
relation and assigned to the respective \textit{LinkedGeoData} class. In addition, the \corpus{} knowledge graph provides the entity's latitude and longitude.
\item \voc{prov}{Entity}: For tracking the origins of an embedding, each geographic entity is linked to the dataset it is extracted from (\voc{prov}{Collection}). Through versioning of these datasets, the \corpus{} corpus and the \corpus{} knowledge graph can be extended in future versions. 
\item \voc{geo}{EmbeddedSpatialThing}: The geographic entities are linked to other resources representing the same (\voc{owl}{sameAs}) or a related resource (\voc{dcterms}{related}) in Wikidata, DBpedia and Wikipedia. 
\end{itemize}

Listing \ref{berlin_kg} presents the triples describing the geographic entity representing the city of Berlin. These triples provide the geolocation of Berlin, references to its counterparts in Wikidata, DBpedia, Wikipedia and OpenStreetMap, as well as provenance information (the embeddings were extracted from an OSM snapshot from November 2020). Access to the \gvtags{} and \gvnle{} embedding is enabled through the Zenodo DOIs \href{https://doi.org/10.5281/zenodo.4321406}{\texttt{10.5281/zenodo.4321406}} and  \href{https://doi.org/10.5281/zenodo.4957746}{\texttt{10.5281/zenodo.4957746}} pointing to the \gvtags{} and \gvnle{} embeddings, the entity type (\voc{lgd}{Node}) and its identifier ($240109189$).

\begin{lstlisting}[basicstyle=\scriptsize\ttfamily, caption={RDF representation of Berlin in the GeoVectors Knowledge Graph.}, label=berlin_kg] 

geovec:v2_n_240109189 a geovec-s:EmbeddedSpatialThing;
  a lgd:Node;
  geo:longitude "13.3888599"^^xsd:double;
  geo:latitude "52.5170365"^^xsd:double;
  dcterms:identifier 240109189;
  rdfs:label "Berlin";
  dcterms:isPartOf <https://doi.org/10.5281/zenodo.4321406> ;
  dcterms:isPartOf <https://doi.org/10.5281/zenodo.4323008> ;
  owl:sameAs <https://www.wikidata.org/wiki/Q64>;
  dcterms:related
    <https://de.wikipedia.org/wiki/Berlin>;
  dcterms:related
    <http://de.dbpedia.org/resource/Berlin >;
  prov:hadPrimarySource
    <https://www.openstreetmap.org/node/240109189>;
  prov:wasDerivedFrom geovec:v2/collection.
geovec:v2/collection a prov:Collection;
  prov:generatedAtTime "2020-11-10"^^xsd:date;
  owl:versionInfo "1.0".
\end{lstlisting}

%% file: 4_Characteristics.tex
\section{\corpus Embedding Characteristics}
\label{sec:chararcteristics}

\begin{figure*}
    \centering
     \hfill  

    \begin{minipage}[t]{0.8\textwidth}
        \begin{minipage}[t]{0.75\textwidth}
        \begin{tikzpicture}
        \node[inner sep=0pt] (picture) {    
        \includegraphics[width=\textwidth]{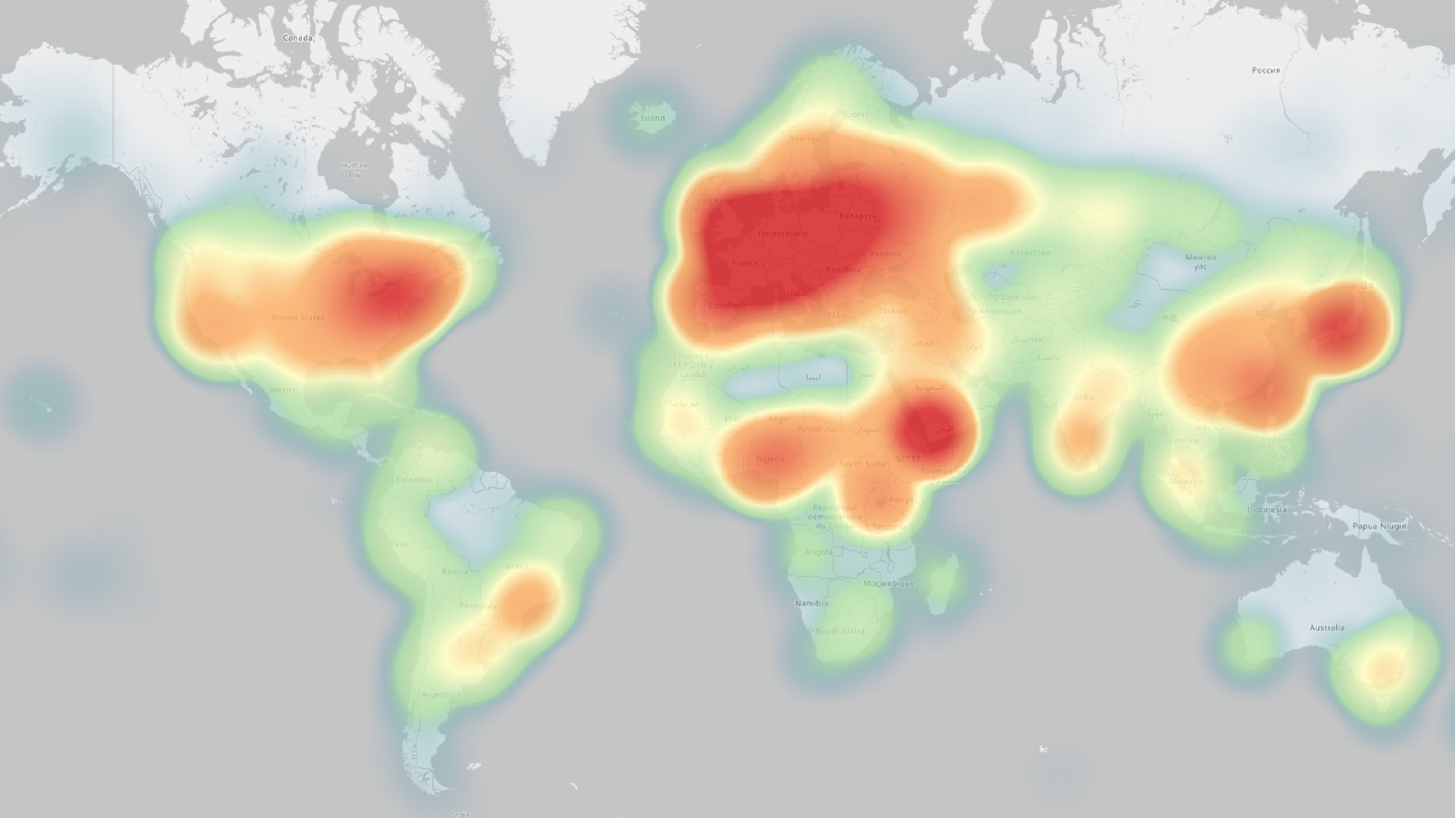}};
    \end{tikzpicture}%
    \end{minipage}    
    \begin{minipage}[t]{0.15\textwidth}
    \includegraphics[width=0.7\textwidth]{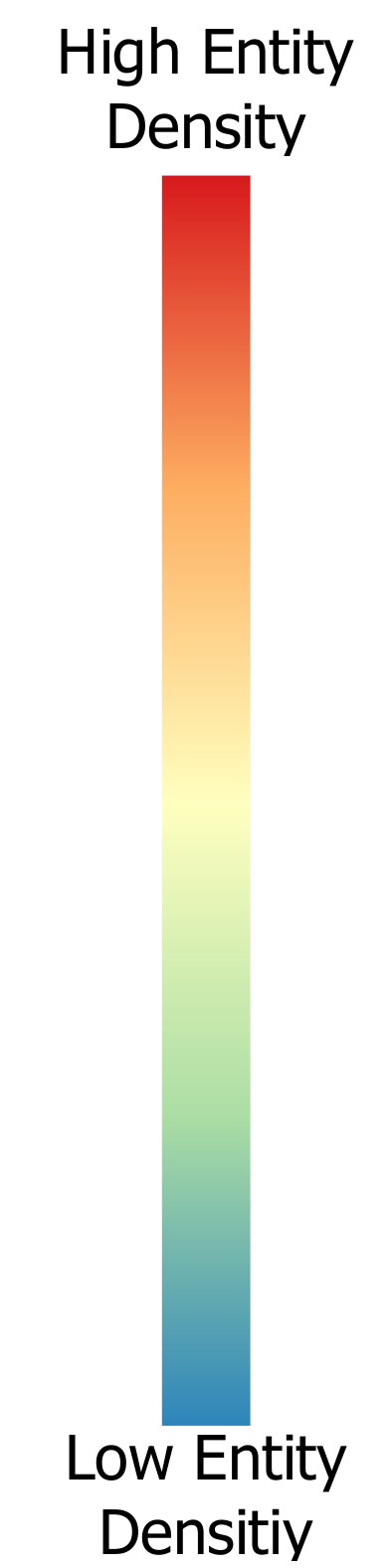}
    \end{minipage} 
    \end{minipage}    
    \hfill  
    \caption{Heatmap visualization of geographic embedding coverage. Map image: \textcopyright OpenStreetMap contributors, ODbL.}
    \label{fig:heatmap}
\end{figure*}

In \corpus V1.0, we extracted representations of nodes, ways, and relations from  OpenStreetMap snapshots at country-level from October 2020\footnotemark[11]. 
We capture all OSM entities having at least one tag. 
Entities without any tags typically represent geometric primitives that isolated carry no semantics. Compound OSM entities such as ways and relations typically subsume such geometric primitives and are better suited for the representation.
Table \ref{tab:statistics_continents} summarizes the number of extracted representations regarding their geographic origin.
In addition, Figure \ref{fig:heatmap} provides a visualization of the geographic coverage of the \corpus corpus.
Overall, we observe high geographic coverage.
In total, \corpus contains representations of over 980 million OpenStreetMap entities. 

The most significant fraction of extracted representations is located in Europe (430 million), followed by North America (240 million) and Asia (150 million).
The number of representations per region follows the distribution of available volunteered information in OpenStreetMap, most prominent in the regions mentioned above. 
Nevertheless, \corpus provides a considerable amount of entity representations for the remaining regions, e.g., 97 million entities for Africa.
We believe that this amount of data is sufficient for many real-world applications.
We expect that the amount of data will further increase in future OSM versions. 
We will include this data in future \corpus releases.

\begin{table}[t!]
    \centering
    \caption{Number of OSM entities contained in \corpus by region.}
    \begin{tabular*}{0.46\textwidth}{p{0.13\textwidth}>{\centering}
    p{0.0525\textwidth}>{\centering}
    p{0.0525\textwidth}>{\centering}
    p{0.065\textwidth}>{\centering\arraybackslash}
    p{0.06\textwidth}}
        \toprule
         Continent & No. Nodes & No. Ways & No.\\Relations  & Total \\
         \midrule
         Africa & $ 9.6 \cdot 10^6 $ & $ 8.7 \cdot 10^7 $ & $ 2.4 \cdot 10^5 $ & $ 9.7 \cdot 10^7 $\\
         Antarctica  & $ 6.9 \cdot 10^3 $ & $ 8.4 \cdot 10^4 $ & $ 9.2 \cdot 10^3 $ & $ 1.0 \cdot 10^5 $\\
         Asia        & $ 1.5 \cdot 10^7 $ & $ 1.8 \cdot 10^5 $ & $ 6.7 \cdot 10^5 $ & $ 1.5 \cdot 10^8 $\\
         Australia/Oceania  & $ 5.2 \cdot 10^6 $ & $  7.6 \cdot 10^6 $ & $ 1.7 \cdot 10^5 $ & $ 1.3 \cdot 10^7 $\\
         Europe        & $ 9.6 \cdot 10^7 $ & $ 3.2 \cdot 10^8 $ & $ 5.6 \cdot 10^6 $ & $ 4.3 \cdot  10^8 $\\
         Central-America & $ 4.4 \cdot 10^5 $ & $ 4.1 \cdot 10^6 $ & $ 1.6 \cdot 10^4 $ & $ 4.6 \cdot 10^6 $\\
         North-America  & $ 5.1 \cdot 10^7 $ & $ 1.9 \cdot 10^8 $ & $ 1.8 \cdot 10^6 $ & $ 2.4 \cdot 10^8 $\\
         South-America  & $ 8.3 \cdot 10^5 $ & $ 2.6 \cdot 10^7 $ & $ 3.9 \cdot 10^5 $ & $ 3.5 \cdot 10^7 $\\
         \midrule
         Total & $ 1.8 \cdot 10^8 $ & $ 7.8 \cdot 10^8 $ & $ 9.1 \cdot 10^6 $ & $ 9.8 \cdot 10^8 $ \\ 
         \bottomrule
    \end{tabular*}
    \label{tab:statistics_continents}
\end{table}

%% file: 6_Case_Study.tex
\section{Case Studies}
\label{sec:case_study}

To illustrate the utility of the \corpus embeddings, we have conducted two case studies dealing with the type assertion and link prediction tasks. These case studies were selected to demonstrate how widely adopted machine learning models can benefit from the \corpus{} embeddings based on semantic and geographic entity similarity.
Other potential use cases include but are not limited to next trip recommendation, geographic information retrieval, or functional region discovery.

In both case studies, we use the same widely adopted classifiers:
The \textsc{Random Forest} model is a standard random forest classifier. We use the implementation provided by the scikit-learn library\footnote{\url{https://scikit-learn.org/}} with the default parameters.
The \textsc{Multilayer Perceptron} model is a simple feed-forward neural network. The hidden network layers have the dimensions [200, 100, 100] and use the ReLu activation function.
The classification layer uses the softmax activation function. 
The network is trained using the Adam optimizer and a categorical cross-entropy loss. We use the default parameters from the Keras API\footnote{\url{https://keras.io/}}.
As the purpose of the case studies is to demonstrate the utility of \corpus, rather than 
achieving the highest possible effectiveness of the models, 
we adopt the default model hyper-parameters without any further optimization.

\subsection{Case Study 1: Type Assertion}
\label{sec:osm-entity-classification}

The goal of this case study is to assign Wikidata classes to OSM entities, which aligns well with the established task of completing type assertions in knowledge graphs \cite{paulheim2017knowledge}. We expect that this case study particularly benefits from the semantic dimension of the OSM entities as captured by the \gvtags{} embeddings.

\emph{Test and training dataset generation:}
To obtain a set of relevant Wikidata classes, we first extract all OSM entities that possess an identity link to Wikidata.
All Wikidata classes that are assigned to at least 10,000 OSM entities are selected for this case study. 
This way, we obtain 32 Wikidata classes, including ``church building''\footnote{\url{https://www.wikidata.org/wiki/Q16970}} and ``street''\footnote{\url{https://www.wikidata.org/wiki/Q79007}}, as well as more fine-grained classes such as ``village of Poland''\footnote{\url{https://www.wikidata.org/wiki/Q3558970}}. 
Finally, we balance the classes by applying random under-sampling and split the data into a training set (80\%, 285k examples) and a test set (20\%, 71k examples).

\emph{Performance}:
Table \ref{tab:perfomance_wikidata} presents the classification performance of the \textsc{Random Forest} and \textsc{Multilayer Perceptron} models using \gvtags{} and \gvnle{} in terms of precision, recall and F1-score. 

As expected, we observe that the \gvtags{} embeddings achieve a better performance than the \gvnle{} embeddings concerning all metrics.
In particular, \gvtags{} achieves an F1-score of 85.95\% and 83.43\% accuracy using the \textsc{Multilayer Perceptron} model.
The \textsc{Random Forest} model using \gvnle{} embeddings reaches an F1-score of 50.17\%. This result can be explained by a few classes such as ``village of Poland'' that are correlated with a location.
The results of this case study confirm that the semantic proximity information is appropriately captured by the \gvtags{} embeddings.

\subsection{Case Study 2: Link Prediction}
\label{sec:country-of-origin-classification}

This case study aims to assign OSM entities to their countries of origin. This task is a typical example of link prediction, where the missing object of an RDF triple is identified \cite{paulheim2017knowledge}.
We expect that this case study particularly benefits from the \gvnle{} embeddings based on geographic proximity.

\emph{Test and training dataset generation:}
To obtain a set of countries, we sample OSM entities from the country-specific snapshots\footnote{Country-specific snapshots are available at \url{https://download.geofabrik.de/}.} as described in Algorithm \ref{alg:Sample} and preserve the origin information.
In analogy to case study 1, we select all countries with at least 10,000 examples and obtain 88 different countries.
Again, we balance the examples by applying random under-sampling and split the data into a training set (80\%, 687k examples) and a test set (20\%, 171k examples).

\emph{Performance}:
Table \ref{tab:perfomance_countries} presents the classification performance of the \textsc{Random Forest} and \textsc{Multilayer Perceptron} models using \gvtags{} and \gvnle{} in terms of precision, recall, and F1-score. 
As expected, we observe that \gvnle{} achieves a better performance than the \gvtags{} embeddings concerning all metrics on this task.
In particular, the \gvnle{} embeddings achieve an F1-score of 96.03\% and 94.80\% accuracy using the \textsc{Multilayer Perceptron} classification model.
In contrast, the \gvtags{} embeddings achieve an F1-score of only 29.91\% and 20.20\% accuracy on this task because the OSM tags of an OSM entity are rarely related to its country of origin.
The results of this case study confirm that the \gvnle{} embeddings appropriately capture geographic proximity.

\setlength{\tabcolsep}{0.5em}
\begin{table*}[t!]
    \centering
    \caption{Precision, recall and F1-score (macro averages) and accuracy [\%] of type assertion.}
    \begin{tabular}{lcccccccc}
         \toprule
        & \multicolumn{4}{c}{\gvtags{}} 
		& \multicolumn{4}{c}{\gvnle{}}  \\
		\cmidrule(l{0pt}r{6pt}){2-5} \cmidrule(l{0pt}r{6pt}){6-9} 
		    &  Precision &  Recall & F1 & Accuracy
		    &  Precision &  Recall & F1 & Accuracy\\
         \midrule
         \textsc{Random Forest} & 92.80 & 69.07 & 77.37 & 69.06
         & 70.97 & 41.53 & 50.17 & 41.53\\
         \textsc{Multilayer Perceptron} & 90.18 & 83.41 & 85.95 & 83.43
         & 63.70 & 36.68 & 41.69 & 36.66\\
         \bottomrule
    \end{tabular}
    \label{tab:perfomance_wikidata}
\end{table*}

\begin{table*}[t!]
    \centering
    \caption{Precision, recall and F1-score (macro averages)   and accuracy [\%] of link prediction.}
    \begin{tabular}{lcccccccc}
         \toprule
        & \multicolumn{4}{c}{\gvtags{}} 
		& \multicolumn{4}{c}{\gvnle{}}  \\
		\cmidrule(l{0pt}r{6pt}){2-5} \cmidrule(l{0pt}r{6pt}){6-9} 
		    &  Precision &  Recall & F1 & Accuracy
		    &  Precision &  Recall & F1 & Accuracy\\
         \midrule
         \textsc{Random Forest} & 84.38 & 20.25 & 29.91 & 20.28 
        & 99.08 & 89.79 & 93.67 & 89.78 \\
         \textsc{Multilayer Perceptron}  & 86.68 & 17.21 & 25.39 & 17.23
         & 96.03 & 94.89 & 95.39 & 94.89\\
         \bottomrule
    \end{tabular}
    \label{tab:perfomance_countries}
\end{table*}

%% file: 7_Availability.tex
\section{Availability \& Utility}
\label{sec:avail}

The \corpus{} website\footnote{\url{http://geovectors.l3s.uni-hannover.de/}} provides a dataset description, the embedding framework as well as pointers to the following resources:

\begin{itemize}
    \item \corpus{} embeddings: We provide permanent access to the \corpus{} embeddings and the trained models on Zenodo under the Open Database License\footnote{\url{https://opendatacommons.org/licenses/odbl/}}. To facilitate efficient reuse, we provide embeddings in a lightweight \textit{TSV} format.
    \item The \corpus knowledge graph described in Section \ref{sec:kg} can be queried through a public SPARQL endpoint that is integrated into the \corpus{} website\footnotemark[4]. 
    In addition, we provide an interface for the label-based search of knowledge graph resources. 
    The resources can be accessed both via HTML pages and via machine-readable formats. A machine-readable VoID description of the dataset is provided and integrated into the knowledge graph. New dataset releases will imply knowledge graph updates, where each release is accompanied by a new instance of \voc{prov}{Collection}.
    \item The \corpus{} embedding generation framework presented in Section \ref{sec:generation} is available as open-source software on GitHub\footnotemark[2] under the MIT License.
\end{itemize}

In Section \ref{sec:intro}, we have presented the benefits of using geographic embeddings in a variety of domains \cite{LIU2018183,Xie:2016:LGP:2983323.2983711,Wang:2017:RRL:3132847.3133006}. With \corpus{}, we aim at providing access to easily reusable embeddings of geographic entities that can directly support tasks in these and other domains. Due to the task-independent nature of our embedding generation framework, we envision high generalizability of \corpus{} in a variety of application scenarios.

For demonstrating the effectiveness of the \corpus{} embeddings in different scenarios, we have conducted two case studies presented in Section \ref{sec:case_study}, which illustrate that the \corpus embeddings adequately capture both the semantic and geographic similarity of OSM entities. Therefore, we believe that \corpus{} eases the use of OSM data and is of potential use for many machine learning and semantic applications that rely on geographic data.
Finally, the \corpus{} framework can be reused to infer embeddings from arbitrary OSM snapshots.

For sustainability and compliance with up-to-date OSM data, we plan yearly releases of new embeddings versions.

%% file: 8_RelatedWork.tex
\section{Related Work}
\label{sec:related_work}
This section discusses related work in the areas of geographic embeddings, word embeddings, and knowledge graph embeddings.

\emph{Geographic Embeddings:}
Recently, several algorithms for the creation of domain-specific geographic embeddings emerged.
Popular application domains include location recommendation \cite{LIU2018183,Xie:2016:LGP:2983323.2983711}, human mobility prediction \cite{Wang:2017:RRL:3132847.3133006}, and location-aware image classification \cite{Aodha_2019_ICCV, 9022254}.
In contrast, we provide a corpus of domain-independent embeddings extracted from OpenStreetMap without the supervision of any specific downstream task. 
We believe that a wide range of applications can benefit from the \corpus embeddings.
Mai et al. proposed a location-aware knowledge graph embedding algorithm for question answering \cite{https://doi.org/10.1111/tgis.12629}.
However, the authors use a planar geographic projection that is only capable of capturing specific regions. In contrast, the \corpus corpus covers the whole globe and captures the spherical earth surface.

The neural location embeddings (NLE) \cite{10.1007/978-3-319-68204-4_14} were initially proposed to encode entities included in the GeoNames knowledge base.
We adopt NLE to capture the geographic dimension of OSM data.
In our previous work \cite{TEMPELMEIER2021349}, we introduced an embedding algorithm for OSM that learns embeddings of individual entities.
However, learning a latent representation for each entity on a world scale is not feasible.
Therefore, we chose tag and location-based, effective, and efficient algorithms to infer the \corpus representations.

\emph{Word Embeddings:}
A multitude of natural language processing algorithms adopts word embedding models for downstream tasks. \cite{DBLP:journals/computing/WangZJ20} conducted a recent survey on neural word embeddings algorithms. 
Recent approaches like BERT \cite{devlin-etal-2019-bert}, and ELMo \cite{Peters:2018} exploit the context information, e.g., the word order in sentences, to infer latent representations.
In contrast, the fastText algorithm \cite{joulin-etal-2017-bag} infers the latent representation of each word individually. 
As OSM tags describing geographic entities neither have any natural order nor form any sentences, we choose fastText over BERT and ELMo to create embeddings.

\emph{Knowledge Graph Embeddings:}
Knowledge graph embeddings have recently evolved as an important area to facilitate latent representations of entities and their relations \cite{8047276}. 
General-purpose knowledge graphs like Wikidata \cite{Vrandecic:2014}, DBpedia \cite{10.1007/978-3-540-76298-0_52}, and YAGO \cite{10.1007/978-3-030-49461-2_34}, and even specialized KGs like EventKG \cite{gottschalk2019eventkg} and LinkedGeoData \cite{10.5555/2590208.2590210} typically only include the most prominent geographic entities.
Compared to OpenStreetMap, the number of geographic entities captured in such knowledge graphs is relatively low \cite{TEMPELMEIER2021349}.
For instance, as of June 2021, Wikidata contained less than 8.5 million entities with geographic coordinates, while OpenStreetMap contained more than 7 billion entities.
The specific geographic entities or entity types, e.g., roads or shops,  might not be relevant or prominent enough to be captured by the general-purpose knowledge graphs. Nevertheless, these entities play an essential role for various downstream applications, for instance, for land use classification \cite{SCHULTZ2017206} or in the prediction of mobility behavior \cite{10.1371/journal.pone.0188735}.
Consequently, pre-trained embeddings of popular knowledge graphs, such as Wikidata \cite{DBLP:conf/mlsys/LererWSLWBP19} or DBpedia, lack coverage of geographic entities required by spatio-temporal analytics applications.
In contrast, the \corpus embeddings proposed in this work specifically target geographic entities and ensure adequate coverage in the resulting dataset.

%% file: 9_Conclusion.tex
\section{Conclusion}
\label{sec:conclusion}
In this paper, we presented \corpus{} -- a linked open corpus of OpenStreetMap embeddings.
\corpus contains embeddings of over 980 million OpenStreetMap entities in 180 countries that capture both their semantic and geographic entity similarity. 
\corpus constitutes a unique resource of geographic entities concerning its scale and its latent representations.
The \corpus knowledge graph provides a semantic description of the corpus and includes identity links to Wikidata and DBpedia.
We further provide an open-source implementation of the proposed \corpus embedding framework that enables the dynamic encoding of up-to-date OpenStreetMap snapshots for specific geographic regions.